\documentclass[letterpaper, 10 pt, conference]{ieeeconf}  

\IEEEoverridecommandlockouts                              

\usepackage{amsmath,amssymb,amsfonts}
\usepackage{graphicx}

\usepackage{newtxtext} 
\usepackage[dvipsnames]{xcolor}
\usepackage{booktabs}
\usepackage{multirow}

\usepackage{threeparttable}

\newcommand{\sparse}{r_{\mathrm{sparse}}}
\newcommand{\optimal}{r_{\mathrm{optimal}}}
\newcommand{\mmin}{\mathrm{tol}}
\newcommand{\thresh}{\mathrm{t}_{\mathrm{req}}}
\newcommand{\pgoal}{p_\mathrm{goal}}
\newcommand{\ptol}{p_\mathrm{tol}}

\newcommand{\update}[1]{#1}

\definecolor{mygreen}{RGB}{89, 182, 91}
\definecolor{darkgreen}{RGB}{63, 128, 70}

\newcommand{\bt}[1]{\textcolor{mygreen}{#1}}
\newcommand{\bbt}[1]{\textcolor{darkgreen}{$\mathbf{#1}$}}
\newcommand{\mc}[3]{\multicolumn{#1}{#2}{#3}}
\title{\LARGE \bf
Generating Realistic Arm Movements in Reinforcement Learning: A Quantitative Comparison of Reward Terms and Task Requirements
}

\author{Jhon P.F. Charaja$^{* 1}$, Isabell Wochner$^{2}$, Pierre Schumacher$^{1,3}$, Winfried Ilg$^{1}$, Martin Giese$^{1}$,\\ Christophe Maufroy$^{4,5}$, Andreas Bulling$^{6}$, Syn Schmitt$^{7}$, Georg Martius$^{3,8}$, and Daniel F.B. Haeufle$^{1,2}$
\thanks{$^{*}$corresponding author: jhon.charaja-casas@uni-tuebingen.de, 
$^{1}$Hertie Institute for Clinical Brain Research, and Centre for Integrative Neuroscience, University of Tübingen, Germany. %
$^{2}$Institute of Computer Engineering (ZITI), Heidelberg University, Germany. %
$^{3}$Max Planck Institute for Intelligent Systems, Tübingen, Germany. %
$^{4}$Institut for Industrial Manufacturing and Management (IFF), University of Stuttgart, Germany. %
$^{5}$Fraunhofer Institute for Manufacturing Engineering and Automation (IPA), Germany.%
$^{6}$Institute for Visualization and Interactive Systems, University of Stuttgart, Germany. %
$^{7}$Institute for Modelling and Simulation of Biomechanical Systems, University of Stuttgart, Germany. %
$^{8}$Department of Computer Science, University of Tübingen, Germany.
}
\thanks{This work was financed by the Baden-W\"{u}rttemberg Stiftung in the scope of the AUTONOMOUS ROBOTICS project \textit{iAssistADL} granted to WI, MG, SS, and DH. The authors thank the International Max Planck Research School for Intelligent Systems (IMPRS-IS) for supporting Jhon Charaja.}%
}

\begin{document}
\maketitle
\thispagestyle{empty}
\pagestyle{empty}

\begin{abstract}
Mimicking of human-like arm movement characteristics involves considering three factors during control policy synthesis: (a) task requirements, (b) noise during movement execution, and (c) optimality principles. Previous studies showed that when these factors (a-c) are considered individually, it is possible to synthesize arm movements that either kinematically match experimental data or reproduce the stereotypical triphasic muscle activation pattern. However, no quantitative comparison has assessed the realism of arm movements generated by each factor, nor has it been determined whether combining these factors results in movements with human-like kinematic characteristics and the triphasic muscle pattern. To investigate this, we used reinforcement learning to learn a control policy for a musculoskeletal arm model, aiming to discern which combination of factors (a-c) results in realistic arm movements according to four frequently reported stereotypical characteristics. Our findings indicate that incorporating velocity and acceleration requirements into the reaching task, employing reward terms that minimize mechanical work, hand jerk, and control effort, along with the inclusion of noise during movement, leads to realistic human arm movements by reinforcement learning. We expect that the gained insights will help in the future to better predict desired arm movements and corrective forces in wearable assistive devices.
\end{abstract}

\section{Introduction}
In aging societies, the number of people benefiting from motor rehabilitation is on the rise~\cite{cieza2020global}. Assistive devices promise support in activities of daily living, e.g., reaching for tools and objects~\cite{maciejasz2014survey}. The design and control of assistive devices would benefit from models accurately predicting human movement. Reinforcement learning in combination with biomechanical models can lead to the emergence of natural characteristics, such as gait kinematics~\cite{schumacher2023natural} and hand trajectories~\cite{fischer2021reinforcement}. However, this requires identifying reward terms and task requirements that lead to realistic movements.

Arm-reaching movements exhibit highly stereotypical kinematics and temporal characteristics. Important characteristics documented in literature are: (i)~roughly straight hand trajectories, (ii)~bell-shaped tangential velocity profiles \cite{morasso1981spatial, abend1982human}, (iii)~triphasic muscle activation pattern~\cite{wierzbicka1986role,kistemaker2006equilibrium}, i.e., the alternating activation of agonist and antagonist muscles, and (iv)~linear relationship between movement time~(MT) and index of difficulty~(ID) (a.k.a Fitts's law)~\cite{mackenzie1989note}. Several optimality principles have been proposed for deterministic prediction of arm-reaching movements, such as minimal work, jerk or muscular effort~\cite{berret2011evidence, flash1985coordination}. Flash et al. \cite{flash1985coordination} found that minimization of hand jerk predicts characteristics (i\&ii) in point-to-point movements. Wochner et al.~\cite{wochner2020optimality} indicated that minimization of mechanical work, jerk, and muscle stimulation command~(effort) predicts characteristics~(i\&ii) in point-to-manifold movements. Finally, Ueyama et al.~\cite{ueyama2021costs} demonstrated that minimization of control effort and consideration of position, velocity, and force requirements in the reaching task predict characteristics (i-iii). As a stochastic approach, Fischer et al.~\cite{fischer2021reinforcement} applied constant and signal-dependent noise of muscle stimulation amplitude. Combined with minimization of movement time they were able to reproduce characteristics (i\&ii\&iv) on point-to-point movements. To our knowledge, no simulation approach has investigated all four characteristics.

More precisely, three factors influence the resulting behavior of the control policy to generate human characteristics of arm movement: (a)~the chosen task requirements, (b)~inclusion of noise during movement execution and (c)~the chosen optimality principles. Some of these factors have been evaluated based on their ability to generate kinematic characteristics that match experimental data, while others evaluated the emergence of the triphasic muscle activation pattern. However, no quantitative comparison has been conducted on the realism of the arm movement generated by each factor; as well as whether a partial or total combination of all factors results in arm movements with human-like kinematic and muscle activation pattern.

The purpose of this study is to investigate which combination of factors (a-c) result in realistic arm movements according to the four stereotypical characteristics (i-iv) defined above. We test this using reinforcement learning to learn a control policy for a musculoskeletal arm model and systemically investigate a combination of (a) the chosen task requirements, (b) inclusion of noise during movement execution and (c) the chosen optimality principles with the aim of methodically evaluating their contribution---for the first time---in one model. We expect that the gained insights will help in the future to better predict desired movements and corrective forces in assistive devices.

\begin{figure*}
    \centering
    \includegraphics{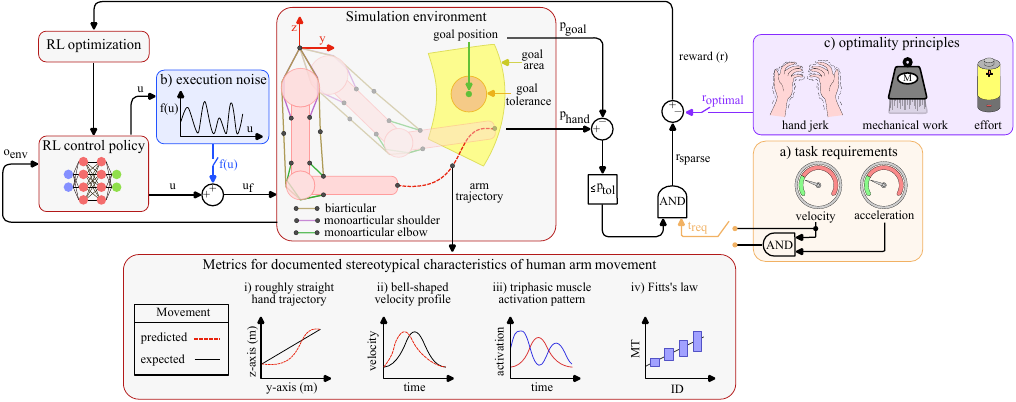}
    \caption{Framework to systematically combine three factors that generate arm movements: (a) different task requirements, (b) inclusion of noise during movement execution, and (c) optimality principles grounded on the minimization of mechanical work, hand jerk and muscle stimulation command (effort). Each combination creates a unique learning environment with distinctive challenges and movement priorities: execution noise modifies the control commands, while optimality principles and additional task requirements shape the reward. Shown below are the metrics for documented stereotypical characteristics of human arm movement: (i) roughly straight hand trajectory, (ii) bell-shaped velocity profile, (iii) triphasic muscle activation pattern, and (iv) Fitts's law.}
    \label{fig:main_framework}
\end{figure*}

\section{Methods}
In a nutshell, the factors (a-c) are categorized into two primary domains: models and task requirements. We investigate four models: the \textit{baseline} model only aims at minimizing movement time. In the other three models, the \textit{baseline} model is combined with either \textit{execution noise} (b), \textit{optimality principles} (c) characterized by the minimization of mechanical work, hand jerk, and muscle stimulation commands, or a \textit{hybrid} model that considers execution noise and optimality principles. For the task requirements~(a) we consider three potential configurations: position only (pos), position and velocity (pos-vel), and position, velocity, and acceleration (pos-vel-acc), all aiming to fulfill respective kinematic constraints at the target location. Details will be given below. This organization facilitates the exploration of how different combinations of each factor (a-c) influence the behavior of the resulting control policy, as is illustrated in Figure~\ref{fig:main_framework}. By finally analyzing the resulting movements according to the stereotypical characteristic (i-iv) of human arm movement, we can identify essential elements for generating arm movements that exhibit human characteristics without enforcing them explicitly as reward terms.

The simulation workflow requires: generation of muscle stimulation commands, simulation of human arm dynamics, calculating rewards, and using metrics for goal-oriented movements. In the following subsections, these will be described in detail.

\subsection{Muscle stimulation commands}
The RL agent utilizes Maximum a Posteriori Optimization~(MPO)~\cite{abdolmaleki2018maximum} combined with DEP-RL~\cite{schumacher2022dep} for exploration; a novel approach that demonstrated robust performance controlling musculoskeletal systems. The MPO implementation follows the default settings provided by the TonicRL library~\cite{pardo2020tonic}\footnotemark{}, and the DEP-RL configuration mirrors the hyperparameters outlined for the same arm model in~\cite{schumacher2022dep}.

The RL agent undergoes training with the inclusion of execution noise (when activated) and random position targets sampled from an area determined by the arm model kinematics. The control policy $(\pi)$ computes muscle stimulation commands ($u$) based on the current observation from the environment ($o_\mathrm{env}$). The execution noise is introduced as $u_f = (1 + \eta_1) u + \eta_2$, modifying the amplitude of control commands $u$, where $\eta_{1}$ represents the signal-dependent noise, $\eta_{2}$ represents the constant motor noise and $u_f$ is the applied muscle stimulation command. Both noise signals are random Gaussian variables, each with a mean of $0$. The standard deviation for $\eta_1$ is $0.103$ and for $\eta_2$ is $0.185$ \cite{van2004role}.

\footnotetext{except for the following parameters: batch size with $256$, batch iteration with $30$, steps before batches with $1e6$ and steps between batches with $50$.}

\subsection{Simulation of human arm dynamics}
The physics engine MuJoCo \cite{todorov2012mujoco} simulates the muscle activation dynamics resulting in the generation of muscle forces that drive the arm movements. In MuJoCo, a musculoskeletal model of the human arm with two degrees of freedom and six actuated muscles is available \cite{todorov2012mujoco}. The original model was modified to generate arm movements in the sagittal plane (considering gravity). The position error is calculated between the tip of the forearm and the desired position (reaching goal). The RL environment considers the same initial conditions of the arm model for all episodes: $0^\circ$ for the shoulder angle, $90^\circ$ for the elbow angle, zero joint velocities and zero muscle activation level. The environment observation comprises Cartesian states\footnotemark{}, joint states\footnotemark{}, muscle states\footnotemark{}, mechanical work, hand jerk and the goal position. The agent's policy network generates control commands every $10$\,ms while the MuJoCo physics engine updates the arm model every $2$\,ms and uses the same control command for five consecutive time steps.
\footnotetext[2]{position, velocity and acceleration of the hand, i.e., tip of the forearm.}
\footnotetext[3]{position, velocity, acceleration and jerk of each arm joint}
\footnotetext[4]{muscle activity, muscle forces, muscle lengths and muscle velocities}

\subsection{Reward formulation}
Previous studies have successfully generated reaching arm movements utilizing an optimal control framework \cite{ueyama2021costs, wochner2020optimality}. This methodology incorporates a terminal cost that penalizes deviations from a desired final state and an accumulated cost associated with the states and control commands during the trajectory \cite{jonsson2010optimal}. Building upon this, the reward function consists of a sparse reward linked to the fulfillment of kinematic requirements at the end of the trajectory and a dense reward associated with executing optimal movements.

The immediate reward function is a combination of
\begin{equation}
    r = c_1 \sparse - c_2 \optimal,
\end{equation}
where $c_{1,2}$ represent weighting coefficients to establish priority during the generation of arm movements, $\sparse$ penalizes movement duration and $\optimal$ encourages optimal behavior based on the optimality principles described above. We select: $c_1=0.2$ and $c_2=0.8$.

Generally, arm-pointing movements are executed quickly. Consequently, previous research employed a constant negative reward for each step transition until the position requirement is met~\cite{fischer2021reinforcement}. Additionally, Ueyama et al.~\cite{ueyama2021costs} found that velocity and force requirements influence the stereotypical triphasic activation pattern. Considering these findings, we incorporate terminal velocity and acceleration (proportional to force in Cartesian space) requirements into the reaching task. Therefore, $\sparse$ depends on meeting the goal tolerance in position as well as the additional kinematic requirements.
\begin{equation}
  r_{\mathrm{sparse}}=\left\{
  \begin{array}{@{}ll@{}}
     0, & \text{if}\ \|p_\mathrm{hand}-\pgoal \| \leq \ptol \And \thresh=\text{true}\\
    -1, & \text{otherwise},
  \end{array}\right.
\end{equation} 
where $p_\mathrm{hand}$ is the Cartesian hand position, $\pgoal$ is the Cartesian desired position, $\ptol$ represents goal tolerance and $\thresh$ is the state of the additional task requirements as
\begin{equation*}
  \thresh=\left\{
  \begin{array}{@{}ll@{}}
     \text{true}, & \text{pos}\  \\
     \|v\| \leq v_\mmin, & \text{pos-vel}\ \\
     \|v\| \leq v_\mmin \And \|a\| \leq a_\mmin, & \text{pos-vel-acc}\\
  \end{array}\right.
\end{equation*}
where $v, a$ are hand velocity and acceleration, $v_\mmin, a_\mmin$ are tolerance for velocity and acceleration. We choose $v_\mmin$, $a_\mmin$ to be $10\%$ of the maximum values observed solely under position task requirement: $v_\mmin=20\,\frac{\mathrm{cm}}{\mathrm{s}}$ and $a_\mmin=100\,\frac{\mathrm{cm}}{\mathrm{s}^2}$.

Furthermore, we consider four values for $\ptol$ to address various difficulty levels in the reaching task. The difficulty associated with reaching movements can be calculated using the index of difficulty~($\mathrm{ID}$)~\cite{mackenzie1989note}, defined as $\mathrm{log}_2 \left(\frac{D}{W} + 1 \right)$, where $D$ represents goal distance and $W=2\ptol$ represents endpoint variability. The values are selected conveniently to ensure that the resulting difficulty indices ($\mathrm{ID}=2$ to $5$) are integers: $D=63$\,cm (used for evaluation) and $\ptol= 10.5$\,cm, $4.5$\,cm, $2.1$\,cm, $1.0161$\,cm. For each combination of model and task requirements, one RL agent is trained for each tolerance value, resulting in a total of $48$ RL agents.

Exclusively focusing on minimizing movement time will generate bang-bang control solutions with asymmetric velocity profiles \cite{harris1998signal}. Wochner et al. \cite{wochner2020optimality} found that bell-shaped velocity profiles emerge in point-to-manifold tasks only when optimal behavior considers the minimization of mechanical work, hand jerk, and muscle stimulation commands (related to muscular effort). As both Berret et al. \cite{berret2011evidence} and Wochner et al. \cite{wochner2020optimality} suggested that it is crucial to consider a combination of optimality principles to tackle the redundancy problem, we therefore, consider the suggested combination $\optimal$ of three optimality principles as: 
\begin{equation}
    \optimal = \frac{c_3 r_\mathrm{effort} + c_4 r_\mathrm{jerk} +  c_5 r_\mathrm{work}}{c_3 + c_4 + c_5},
\end{equation}
where $c_{3,4,5}$ set priority between optimality principles, $r_\mathrm{effort}$ is computed as mean value of muscle stimulation commands, $r_\mathrm{jerk}$ is estimated by finite difference computation between the current and one previous acceleration values, instantaneous work (power) $r_\mathrm{work}$ is computed as $|\dot{\phi}_1 \tau_1| + |\dot{\phi}_2 \tau_2|$, where $\dot{\phi}_{1,2}$ represents the angular velocity of shoulder and elbow, and $\tau_{1,2}$ indicates the torque of shoulder and elbow. We normalize each optimality principle by its observed maximum value:  $r_\mathrm{jerk}=1000\,\frac{\text{m}}{\text{s}^3}$ and $r_\mathrm{work}=100\,\text{J}$. In pre-tests, we found that smooth muscle profiles were only achieved if all the three terms are considered with the following coefficients: $c_3=1$, $c_4=8$, and $c_5=1$.

\subsection{Metrics for goal-oriented movements}
We evaluate each agent in its training environment. The position target for all agents is positioned $29.5$\,cm to the left and $55.7$\,cm upward relative to the tip of the forelimb. We run $1000$ rollouts for each agent to capture their average behavior. Since each test episode has a different movement time~(MT), we temporally normalize the recorded data to make trajectory characteristics comparable. We recognize outliers by examining the velocity profile. Any trajectory with an integral velocity beyond the interquartile range (between the $25$th and $75$th percentiles) is excluded from consideration. The mean computed over the remaining rollouts is employed for the analysis of the movement's characteristics.

The performance of all trained agents is quantified with four metrics associated with the stereotypical characteristic~(i-iv) observed in goal-oriented movements:
\begin{enumerate}
\renewcommand{\labelenumi}{\roman{enumi}.}

\item \textbf{Straight line deviation ($p_\text{line}$):} This metric reports the R-squared between the straight line from initial point to target and the actual hand trajectory.

\item \textbf{Bell-shaped velocity profile ($v_\text{bell}$):} We determine the onset and offset of the velocity profile by the threshold  $v>0.1v_\text{max}$ of the peak velocity. A Gaussian is fitted between onset and offset of the velocity profile. The Gaussian strictly considers peak velocity as amplitude, and the \texttt{fit} function of $\mathrm{MatLab}$ computes the mean and standard deviation of the Gaussian. This metric~($v_\text{bell}$) reports the R-squared to indicate how bell-shaped each velocity profile is.

\item \textbf{Triphasic muscle pattern ($u_\text{triphasic}$):} This metric analyzes the muscle activation pattern of each agonist-antagonist muscle pair in the arm. The aim is to capture if an antagonistic pair changes operation mode, e.g., if in the beginning elbow flexor is actively flexing the elbow and then, elbow extensor activity rises and elbow flexor activity falls to decelerate the movement, this is considered a second phase. We quantify this by evaluating if muscle activation slopes exchange directions and by the threshold $\Delta > 0.25\, \Delta_\mathrm{max}$ and $\Delta > 1.5\,\mathrm{e}-3$  of difference between them. If this is the case, it is considered a new phase of muscle activation.  The metric verifies if the reported triphasic pattern in the literature \cite{ueyama2021costs} occurs in a muscle pair, assigning a score of $1$ if true and $0$ otherwise.

\item \textbf{Fitts's law ($R_F$))}: This metrics reports the correlation coefficient $R_F$ to indicate how strong the linearity is.
\end{enumerate}

\begin{table*}[!ht]
\centering
\begin{threeparttable}

\caption{Analysis of movement characteristics of each combination of model and task requirement using the proposed metrics\tnote{§}.}
\label{tab:results}
\begin{tabular}{ c c     c c c c |  c c c c |  c c c c | c c c c}
\toprule
 \multirow{4}*{Metric} & \multirow{4}{6em}{\centering Task requirements} & \multicolumn{16}{c}{Models}\\
\cline{3-18}
& & \multicolumn{4}{c}{Baseline} & \multicolumn{4}{|c}{Execution noise} & \multicolumn{4}{|c}{Optimality principles} & \multicolumn{4}{|c}{Hybrid} \\
\cline{3-18}
& & \mc{4}{c}{Index of difficulty (ID)} & \mc{4}{|c}{Index of difficulty (ID)} & \mc{4}{|c}{Index of difficulty (ID)} & \mc{4}{|c}{Index of difficulty (ID)} \\

& & $2$  & $3$ & $4$ & $5$ & $2$  & $3$ & $4$ & $5$ & $2$  & $3$ & $4$ & $5$ & $2$  & $3$ & $4$ & $5$ \\
\cline{1-18}
\multirow{3}*{$p_\mathrm{line}$}  

 & pos         & $0.97$         & $0.97$       & $0.97$     & $\bt{1.0}$
               & $\bt{0.99}$    & \bbt{1.0}    & $0.97$     & $0.96$ 
               & $\bt{0.96}$    & $0.95$       & $0.94$     & $\bt{0.99}$ 
               & $\bt{0.98}$    & $\bt{0.94}$  & $0.96$     & $0.98$ \\
               
 & pos-vel     & $0.91$         & $0.95$        & $0.95$        & $0.97$ 
               & $\bt{0.99}$    & \bbt{1.0}     & $0.99$        & $0.99$ 
               & $\bt{0.96}$    & $0.97$        & $0.93$        & $0.94$ 
               & $0.97$         & $0.92$        & $\bt{0.99}$   & \bbt{1.0} \\
               
 & pos-vel-acc & \bbt{1.0}      & \bt{$0.98$}   & $\bt{0.98}$   & $\bt{1.0}$ 
               & $0.98$         & $0.99$        & \bbt{1.0}     & \bbt{1.0} 
               & $\bt{0.96}$    & $\bt{0.99}$   & $\bt{0.96}$   & $0.98$ 
               & $0.95$         & $0.93$        & $0.97$        & $0.98$ \\[5px]
 
 \multirow{3}*{$v_\mathrm{bell}$}  
 & pos\tnote{\dag}  &  - & - & - & -  &  - & - & - & -  &  - & - & - & -  &  - & - & - & -  \\
 & pos-vel       & $0.86$       & $0.73$        & $0.80$         & $0.94$ 
                 & $\bt{0.81}$  & $\bt{0.80}$   & $0.76$         & $0.90$ 
                 & $0.86$       & $0.88$        & $\bt{0.82}$    & $\bt{0.91}$ 
                 & $0.88$       & $0.83$        & $0.89$         & $0.94$ \\
                 
 & pos-vel-acc   & \bbt{0.90}  & $\bt{0.78}$    & $\bt{0.93}$    & $\bt{0.95}$ 
                 & $0.74$       & $\bt{0.80}$   & $\bt{0.79}$    & $\bt{0.95}$ 
                 & $\bt{0.89}$  & \bbt{0.92}    & $0.81$         & $\bt{0.91}$ 
                 & \bbt{0.90}   & \bbt{0.92}    & \bbt{0.95}     & \bbt{0.97} \\[5px]

 \multirow{3}*{$u_\mathrm{triphasic}$}  
 & pos              & $0$ & $0$ & $1$ & $1$ 
                    & $0$ & $0$ & $0$ & $0$  
                    & $0$ & $0$ & $0$ & $1$  
                    & $0$ & $0$ & $0$ & $0$  \\
 
 & pos-vel          & $1$\tnote{*} & $1$ & $1$ & $1$  
                    & $1$ & $1$ & $1$ & $1$  
                    & $1$ & $1$ & $1$ & $1$  
                    & $1$ & $1$ & $1$ & $1$  \\
 
 & pos-vel-acc      & $1$\tnote{*} & $1$ & $0$ & $1$\tnote{*}  
                    & $1$ & $1$ & $1$ & $1$\tnote{*}  
                    & $1$ & $1$ & $1$ & $0$  
                    & $1$ & $1$ & $1$ & $1$\tnote{*} \\[5px]

\multirow{3}*{$R_F$\tnote{\ddag}}  
& pos           & \mc{4}{c|}{\bt{$0.985$}} 
                & \mc{4}{c|}{$0.967$} 
                & \mc{4}{c|}{$0.974$}  
                & \mc{4}{c}{$0.967$} \\

& pos-vel       & \mc{4}{c|}{$0.956$} 
                & \mc{4}{c|}{\bt{$0.997$}} 
                & \mc{4}{c|}{\bbt{0.998}}  
                & \mc{4}{c}{\bt{$0.983$}} \\

& pos-vel-acc   & \mc{4}{c|}{$0.963$} 
                & \mc{4}{c|}{$0.985$} 
                & \mc{4}{c|}{$0.986$}  
                & \mc{4}{c}{$0.929$} \\[1em]

 \bottomrule
\end{tabular}
\begin{tablenotes}
\item[§] The highest values between task requirements for each metric, model, and difficulty index are highlighted in green. Among these values, the best performance across difficulty indexes for each metric is highlighted in bold dark green.
\item[\dag] The trajectories of this terminal condition are invalid for the $v_\text{bell}$ metric as they do not reach the lower threshold of $10\%$ of the peak velocity.
\item[\ddag] $R_F$ has only value because this metric determines the correlation coefficient across all difficulty indices (ID=2...5).
\item[*] These combinations exhibit two or three muscle pairs with a triphasic muscle pattern.
\end{tablenotes}
\end{threeparttable}
\end{table*}

\section{Results}
Overall, incorporating velocity and acceleration requirements into the reaching task (pos-vel-acc), employing reward terms that minimize mechanical work, hand jerk, and \update{control effort}, along with the inclusion of noise during movement, leads to the most realistic arm movement according to the four proposed metrics (i-iv). Furthermore, increasing index of difficulty, from $\mathrm{ID}=2$ to $5$, yields more bell-shaped velocity profiles and the emergence of the third phase in the muscle activation pattern. These results are presented in Table \ref{tab:results}, which shows the performance of all agents for each proposed metric across all difficulty indices ($\mathrm{ID} = 2 \text{ to } 5$). Note that in Table \ref{tab:results}, $R_F$ only displays one value, as this metric utilizes all difficulty indices to determine how strong the linear relationship (correlation) between movement time (MT) and index of difficulty (ID) is (Fitts's law). Also, the velocity profiles obtained with only position task requirement (pos), do not reach the lower threshold of $10\%$ of the peak velocity; consequently, these velocity profiles are not considered for the $v_\mathrm{bell}$ metric (displayed as solid line "-").

\subsection{Straight line deviation ($p_\mathrm{line}$)}
\begin{figure}[!ht]
\centering
\includegraphics{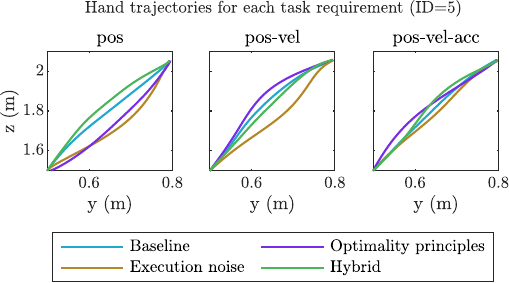}
\caption{Hand trajectories generated by all models, considering the three possible task requirements and difficulty index $\mathrm{ID}=5$.}
\label{fig:pos_profiles}
\end{figure}

The best performance in terms of straight line deviation $p_\mathrm{line}$ across the majority of difficulty indices ($\mathrm{ID}=2$ to $5$) for baseline, execution noise and optimality principles models is linked to pos-vel-acc task requirement (Tab.~\ref{tab:results}). Conversely, the best performances of the hybrid model are distributed across pos and pos-vel task requirements. All hand trajectories with difficulty index $\mathrm{ID}=5$ are illustrated in Figure~\ref{fig:pos_profiles}. The figure illustrates the progressive straightening of hand trajectories as more kinematic requirements are incorporated into the main task. It is noteworthy that even the worst $p_\mathrm{line}$ values ($0.91$, $0.92$, ...) still represent lines that we would consider roughly straight. Consequently, solely relying on the $p_\mathrm{line}$ metric makes it implausible to indicate which combination will yield the most realistic hand trajectory.

\subsection{Bell-shaped velocity profile ($v_\mathrm{bell}$)}
\begin{figure}[!ht]
\centering
\includegraphics{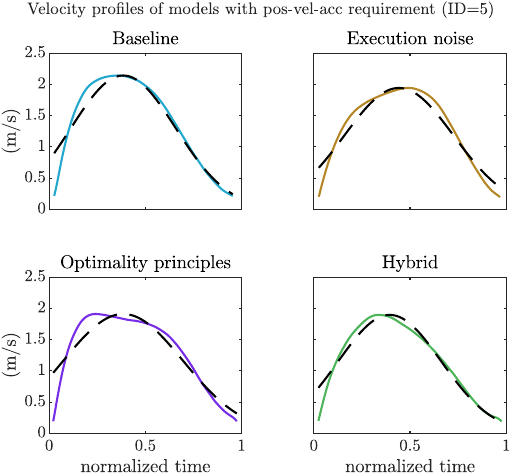}
\caption{Velocity profiles generated by each model, considering velocity and acceleration requirements into main task and difficulty index $\mathrm{ID}=5$. The dashed line represents the fitted Gaussian model.}
\label{fig:vel_profiles}
\end{figure}

The best performance in terms of bell-shaped velocity profile $v_\mathrm{bell}$ across the majority of difficulty indices ($\mathrm{ID}=2$ to $5$) for baseline, execution noise, optimality principles and hybrid models is linked to pos-vel-acc task requirement (Tab.~\ref{tab:results}). \update{The hybrid model combined with pos-vel-acc task requirement, consistently exhibits the highest $v_\mathrm{bell}$ values, i.e., most bell-shaped velocity profiles, across all difficulty indices}. In addition, Table~\ref{tab:results} reveals a increasing trend of $v_\mathrm{bell}$ values with increasing index of difficulty. The velocity profile for $\mathrm{ID}=5$ of each model with pos-vel-acc task requirement are shown in Figure~\ref{fig:vel_profiles}. The figure illustrates that all models align well with the right side of the Gaussian model, and fitting errors arise from the left side.

\subsection{Triphasic muscle pattern ($u_\text{triphasic}$)}
\begin{figure}[!ht]
\centering
\includegraphics{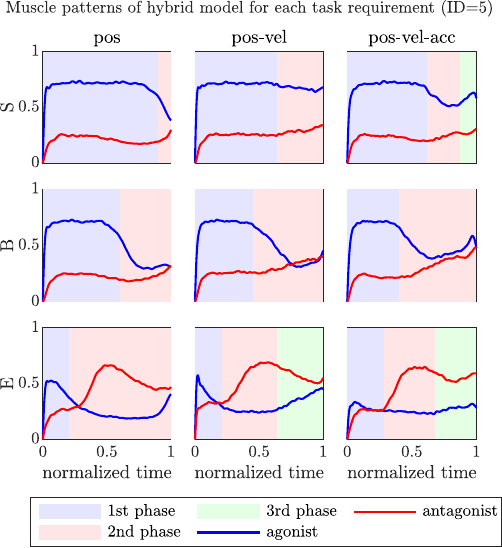}
\caption{Muscle activation pattern of the hybrid model, considering the three possible task requirements. The agonist-antagonist muscle pair of the arm are denoted as: Monoarticular shoulder (S), Biarticular elbow-shoulder (B) and Monoarticular elbow muscle (E). Blue and Red lines represent muscle activation of agonist and antagonist muscles, respectively.}
\label{fig:muscle_patterns}
\end{figure}

\begin{figure}[!htb]
\centering
\includegraphics{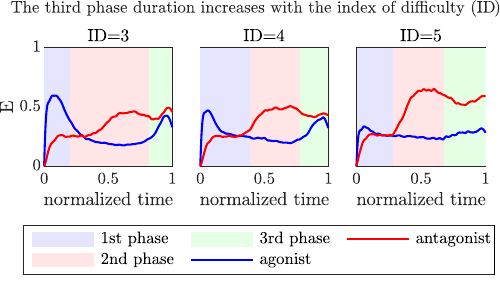}
\caption{Muscle pattern of Monoarticular elbow muscle (E) for hybrid model with pos-vel-acc task requirement. The third phase duration increases with larger index of difficulty (ID), i.e., higher endpoint accuracy.}
\label{fig:ID_muscle}
\end{figure}

The best performance in terms of triphasic muscle pattern $u_\text{triphasic}$ across the majority of difficulty indices ($\mathrm{ID}=2$ to $5$) for all models (Baseline, Execution noise, Optimality principles and Hybrid) is linked to pos-vel and pos-vel-acc task requirements~(Tab. \ref{tab:results}). The muscle patterns for $\mathrm{ID}=5$ of hybrid model for each task requirement are shown in Figure~\ref{fig:muscle_patterns}. The figure illustrates that both pos-vel and pos-vel-acc task requirements give rise to a triphasic muscle pattern in the elbow muscle pair, whereas only position task requirement (pos) results in a biphasic pattern in the three muscle pairs. Furthermore, the figure displays two triphasic muscle patterns for the pos-vel-acc task requirement. Similarly, Figure \ref{fig:ID_muscle} illustrates that duration of the third muscle phase increases with larger index of difficulty (ID).

\subsection{Fitts's law ($R_F$)}
\begin{figure}[!hbt]
\centering
\includegraphics{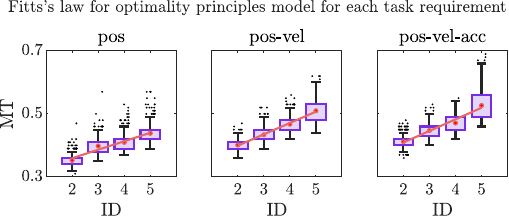}
\caption{Graphic representation of the linear relationship between Movement Time (MT) and Index of Difficulty (ID) for the optimality principles model, considering the three possible task requirements. The red dots represent the average movement time.}
\label{fig:fitt_law}
\end{figure}

The models Execution noise, Optimality principles and Hybrid obtain their highest correlation coefficient $R_F$ when incorporating velocity requirement into the main task (pos-vel) (Tab.~\ref{tab:results}). In contrast, the Baseline model attains its highest correlation coefficient $R_F$ when employing only the position task requirement. The optimality principles model demonstrate slightly superior performance in $R_F$ compared to Baseline, Execution noise and Hybrid models. The linear relationship between Movement Time (MT) and Index of Difficulty (ID) is graphically illustrated in Figure \ref{fig:fitt_law} for the Optimality Principles model considering the three possible task requirements. Additionally, the figure illustrates the increase in movement time variance as more kinematic requirements are incorporated into the main task. It is crucial to emphasize that all combinations generate a robust linear relationship, with correlation coefficients $R_F>0.9$. Consequently, although certain combinations exhibit higher correlation coefficients than others, it is implausible to indicate which combination yields the most realistic arm trajectory based solely on the $R_F$.

\section{Discussion}

Through the systematic combination of factors (a-c), we identify three key considerations for generating human-like characteristics in point-to-point arm movements. First, including both velocity and acceleration as task requirement ((a), pos-vel-acc), results in good or excellent values across all metrics and in the majority of difficulty indices, regardless of the model. Second, using noise b) during movement execution, in combination with reward terms c) minimizing mechanical work, hand jerk, and control effort results in the most bell-shaped velocity profile across the majority of difficulty indexes (except for the position-only task requirement). Third, a higher endpoint accuracy, i.e., a larger index of difficulty (ID), leads to a longer duration of the third muscle phase and velocity profiles with a better-defined bell shape. According to Wierzbicka et al. \cite{wierzbicka1986role}, the role of this third phase is to regulate the braking forces to guide the hand towards the target position. Therefore, the effect of the index of difficulty (ID) can be understood as the prolongation of the deceleration phase to achieve high endpoint accuracy, which in turn smooths the velocity profiles on the right side, enhancing the bell shape. It is noteworthy that although increasing the index of difficulty (ID) has improved the bell shape, larger values will cause the velocity profile to become more positively asymmetric \cite{todorov2004optimality}.

In addition, we found that including the velocity requirement into the reaching task (pos-vel) can yield comparable results to considering both velocity and acceleration (pos-vel-acc). The primary distinction lies in slightly lower values of bell-shaped velocity profile $v_\mathrm{bell}$. Moreover, we found that the triphasic muscle pattern can emerge when incorporating requirements of either velocity~(pos-vel) or velocity and acceleration~(pos-vel-acc) into reaching tasks. This contrasts with Ueyama et al.~\cite{ueyama2021costs}, who suggests position, velocity and force (equivalent to acceleration) are necessary. Unlike Ueyama et al.~\cite{ueyama2021costs}, our approach does not require predefining the movement time for arm movement generation. Although it is not clear how predefining the movement time (MT) influences the emergence of the triphasic muscle pattern, setting a value far from that calculated with Fitts's law could result in unrealistic arm movements.

It is noteworthy to highlight that the metrics $p_\mathrm{line}$, $u_\mathrm{triphasic}$ and $R_F$ do not show large differences across all combinations. This suggests that all investigated models and task requirements (except for position only) lead to somewhat realistic arm movements, at least for the simple planar point-to-point movements investigated here.

Although our control approach generates realistic arm movements with human-like characteristics, our study has some limitations. The arm model used incorporates only two degrees of freedom and six muscles. Consequently, our model does not fully account for the entire joint and muscle redundancy found in a real human arm. Furthermore, the investigated task includes only point-to-point reaching tasks, whereas more openly defined tasks such as point-to-manifold reaching might be interesting for future research, as they offer more freedom in arm movement generation. Previous studies \cite{berret2011evidence, wochner2020optimality} have shown significance differences in the generated arm trajectories using point-to-manifold reaching that have not been observed in point-to-point movements. Moreover, complex movements in a complex arm model may further distinguish between the different combinations such that a solution for predicting realistic human arm movements with RL could aid the development and control of assistive devices.

\bibliographystyle{IEEEtran}
\bibliography{IEEEabrv,mybibfile}
\end{document}